\documentclass[conference]{IEEEtran}
\IEEEoverridecommandlockouts
\usepackage{cite}
\usepackage{amsmath,amssymb,amsfonts}
\usepackage{algorithmic}
\usepackage{graphicx}
\usepackage{textcomp}
\usepackage{xcolor}
\def\BibTeX{{\rm B\kern-.05em{\sc i\kern-.025em b}\kern-.08em
    T\kern-.1667em\lower.7ex\hbox{E}\kern-.125emX}}
\begin{document}

\title{Meta Mask Correction for Nuclei Segmentation in Histopathological Image
\thanks{978-1-6654-0126-5/21/\$31.00 ©2021 IEEE}
}

\author{\IEEEauthorblockN{Jiangbo Shi$^{1,2,3}$, Chang Jia$^{1,2,3}$,
Zeyu Gao$^{1,2,3}$, Tieliang Gong$^{1,2,3}$, Chunbao Wang$^{4}$ and
Chen Li$^{1,2,3^{*}}$}
\IEEEauthorblockA
{$^{1}$National Engineering Lab for Big Data Analytics Xi'an Jiao tong University, Xi'an, Shaanxi 710049, China\\
$^{2}$School of Computer Science and Technology, Xi'an Jiao tong University, Xi'an, Shaanxi 710049, China\\
$^{3}$Shaanxi Province Key Laboratory of Satellite and Terrestrial Network Tech.R\&D,\\Xi'an Jiao tong University, Xi'an, Shaanxi 710049, China\\
$^{4}$Department of Pathology, the First Affiliated Hospital of Xi’an Jiaotong University, Xi'an, Shaanxi 710061, China\\
Email: shijiangbo@stu.xjtu.edu.cn, jc2160500142@gmail.com, betpotti@gmail.com, gongtl@xjtu.edu.cn,\\ bingliziliao2012@163.com, cli@xjtu.edu.cn
}}

\maketitle

\begin{abstract}
Nuclei segmentation is a fundamental task in digital pathology analysis and can be automated by deep learning-based methods. However, the development of such an automated method requires a large amount of data with precisely annotated masks which is hard to obtain. Training with weakly labeled data is a popular solution for reducing the workload of annotation. In this paper, we propose a novel meta-learning-based nuclei segmentation method which follows the label correction paradigm to leverage data with noisy masks. Specifically, we design a fully conventional meta-model that can correct noisy masks using a small amount of clean meta-data. Then the corrected masks can be used to supervise the training of the segmentation model. Meanwhile, a bi-level optimization method is adopted to alternately update the parameters of the main segmentation model and the meta-model in an end-to-end way. Extensive experimental results on two nuclear segmentation datasets show that our method achieves the state-of-the-art result. It even achieves comparable performance with the model training on supervised data in some noisy settings.
\end{abstract}


\section{Introduction}
Nuclei segmentation, which extracts pixel-level mask of each nucleus in the image, is an essential and highly challenging task in digital pathological image analysis. 
Many important downstream tasks, such as genotype-phenotype correlation, survival analysis, etc., rely on the precise segmentation of the nuclei. In recent years, deep neural networks have made great progress in the task of nuclei segmentation \cite{wang2020bending}, \cite{chen2020boundary}, \cite{xie2020integrating}, \cite{shi2019effects}, \cite{wu2021precision}. However, due to the deep layers and many parameters of the DNN, a large amount of high-quality data is usually needed to help the model achieve good performance.
Each pathological image contains tens of thousands of nuclei, and these nuclei show a high-level heterogeneity, such as different shapes, sizes, and chromatin patterns. Moreover, some nuclei appearing in clusters or clumps. All the above situations lead to the difficulty of the fine-grained nuclei annotation, e.g., time-consuming, incorrect, and missing labeling.
Therefore, training a nuclei segmentation model with good performance using noisy annotation is essential and challenging. \par 
Recently, many approaches to medical image segmentation with noisy annotation have been proposed. One popular schema is based on the additional constraints. Qu \cite{qu2020nuclei} et al. generated clustering labels and Voronoi diagrams based on point annotations, supervised the training of Bayesian networks, and selected nuclei with large uncertainties to be labeled by the annotators. Tian \cite{tian2020weakly} et al. supervised model training by iteratively optimizing the generated distance map. This kind of method's main problem is that it needs to generate pseudo-labels based on external knowledge and perform iterative optimization. When there is much noise in the data, it is difficult to converge to better performance. Another popular strategy used to learning with noisy annotation is the loss re-weighting-based method. These works focus on learning a weight matrix to guide the model's optimization direction, thereby reducing the impact of label noise. For example, Spatial reweighting \cite{mirikharaji2019learning} generates importance weights for each pixel based on the pixel-wise loss gradient direction to adjust the contribution of each pixel to model optimization. After that, the MCPM \cite{wang2020meta} method designed a meta-model to protect the segmentation model from the influence of noise annotations. The weight map was generated by inputting the loss value map into the meta-model to illustrate the importance of each pixel, thereby weakening the influence of noise annotation when the model is updated. The limitation of the loss re-weighting-based method is that it can only increase or decrease the weight of the instance contribution in the learning process, and there is a problem of information bottleneck. Because these methods weight the importance of the loss value, the meta-model cannot distinguish the different input pairs if their loss is the same. \par
To solve the above-mentioned problems, we propose a new method Meta Mask Correction termed MMC. We pose the problem as a label correction problem within a meta-learning framework and view the label correction procedure as a meta-process. The idea is to design a fully convolutional meta-model to correct the noise mask and supervise the training of the main segmentation model. Specifically, by inputting the feature maps and noise masks into the meta-model, corresponding corrected masks are generated. The main contributions of this paper include:
\begin{itemize}
    \item Aiming at nuclei segmentation with noisy annotation, based on the label correction schema, we propose a new meta-learning framework to train the segmentation model.
    \item The proposed fully conventional network can learn the mapping relationship between the noise mask and the correction mask under the condition of a small number of clean data.
    \item Extensive experiments were performed on two datasets to verify that the proposed method achieves state-of-the-art performance in nuclei segmentation tasks under different noise types.
\end{itemize}

\section{METHODOLOGY}
\begin{figure}[!t]
\centering
\includegraphics[width=85mm]{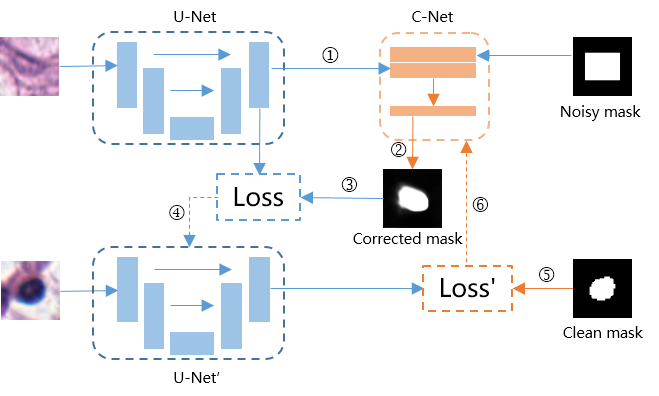}
\caption{MMC Architecture and optimized process. U-Net and C-Net separately represent the main segmentation network and the meta-model.}
\label{f1}
\end{figure}

In this paper, we present a novel method (MMC) to solve the nuclei segmentation task with noise in the histopathological image. When masks of training data contain noise, it is difficult for the segmentation model to achieve satisfactory generalization performance. Noisy masks will provide the wrong direction for updating model parameters, making the model overfit the noise. Simultaneously, directly training a model using a small number of data with fine-grained masks will cause an over-fitting problem, resulting in sub-optimal situations. Zheng et al. \cite{zheng2021meta} proposed a meta-learning-based framework,i.e., MLC, to map noisy labels to clean labels through a multi-layer perceptron under the condition of using only a small amount of clean data. Unlike the image classification model that maps images to various categories, the image segmentation model needs to learn a mapping relationship between images and pixel-level masks. Due to this difference, MLC is hard to handle the segmentation task. Furthermore, MLP can only correct one pixel at a time and does not consider the spatial position relation of different pixels, which is very important for the segmentation task. Inspired by MLC, we designed the correction network structure into a fully conventional network that directly learns a mapping between noisy masks and clean masks. The fully conventional structure can correct an entire mask and consider the spatial position relation of different pixels in an image. The proposed framework and optimization process are shown in Fig. \ref{f1}. The specific calculation process of updating meat-model is as follows: \textcircled{1} Feed image into Current U-Net and compute the logits for prediction, \textcircled{2} Feed noisy mask and prediction logits into C-Net and get its corrected mask, \textcircled{3} Compute the loss with logits and corrected mask, then compute the gradient of the loss with respect to the parameter of the U-Net, \textcircled{4} Update the U-Net parameter while keeping the computation graph for the gradient, \textcircled{5} feed a pair of clean mask and image to the new U-Net and compute the loss, \textcircled{6} Compute the gradient of loss and update the C-Net.

\subsection{Objective functions}
Given a set of clean examples $D=\{x, y\}^m$ and a set of weak annotations data $D'={\{x, y'\}}^M$ in which M and m indicate the number of noisy and clean sample and $m{\ll}M$. $y'$ indicates the noisy mask of image ${x}$. The main model is parameterized as a function with parameter $W$, $y=f_W(x)$. The meta-model is parameterized as a function with parameter $\theta$, $y_c=g_\theta(h(x),y')$ to correct the weak mask $y'$  and example feature $h(x)$ to a more accurate mask. We treat the training of the main segmentation model and meta-model as a bi-level optimization process. Given a fixed meta-model parameters $\theta$,  the optimized solution to $W$ can be found through minimizing the following objective function:
\begin{equation}\label{1}
W_\theta^*=argmin_{\theta}{\mathbb{E}}_{(x,y'){\in}D'}{\mathcal{L}}(f_W(x), g_{\theta}(h(x),y'))
\end{equation}
Given a small number of meta-data with clean mask and optimized $W$, the optimized solution to $\theta$ can be acquired through minimizing the following objective functions:

\begin{equation}\label{2}
{\theta}^*_W=argmin_W{\mathbb{E}}_{(x,y){\in}D}{\mathcal{L}}(y,f_W(x))
\end{equation}

\subsection{Optimization process}
In order to obtain the optimal main and meta-model parameters, we adopt an iterative method to alternately update the values of $W$ and $\theta$ in an end-to-end way. The one-loop optimization algorithm mainly includes the following steps:
\\-Step1: Initialize main segmentation model parameter $W^{(0)}$ and meta-model parameter ${\theta}^{(0)}$.
\\-Step2: For the i-th iteration, the parameters of the segmentation network are temporally updated as in Eq. (\ref{3}), via one step of gradient descent in minimizing the objective function Eq. (\ref{1}). The loss calculation formula of the main segmentation model at time t is shown in Eq. (\ref{4}).
\begin{equation}\label{3}
W_{\theta}^{'(t)} = W^{(t)}-{{\alpha}{\mathbb{E}}_{(x,y^{'}){\in}D^{'}}{\frac{{\partial}L^{(t)}}{{\partial}W}}\bigg|_{W^{(t)}}}
\end{equation}
\begin{equation}\label{4}
{L^{(t)}}={\mathcal{L}}({f_W^{(t)}(x)},g_{\theta}^{(t)}(h(x),y^{'}))
\end{equation}
$\alpha$ is the learning rate of the main segmentation model.
\\-Step3: The meta-model parameter $\theta$ can be updated by minimizing the objective function Eq. (\ref{2}), and the update process is as Eq. (\ref{5}).  
\begin{equation}\label{5}
{\theta}^{(t+1)}={\theta}^{(t)}-{\beta}{{\mathbb{E}}_{(x,y){\in}D}}{{\frac{{\partial}L^{'(t)}}{{\partial}W_{\theta}^{'}}}\bigg|_{W^{'(t)}}}{{\frac{{\partial}W_{\theta}^{'}}{{\partial}{
\theta}}}\bigg|_{{\theta}^{(t)}}}
\end{equation}
where ${L^{'(t)}}={\mathcal{L}}(y,{f_{w^{'}_{\theta}}^{(t)}(x)})$ is calculated on meta-data at time $t+1$. $\beta$ is the learning rate of the meta-model.
\\-Step4: Finally, the main segmentation model parameter $W$ is updated as in Eq. (\ref{6}) by minimizing the objective function Eq. (\ref{1}), loss calculation formula of the main segmentation model at time $t+1$ is as shown in Eq. (\ref{7}). 
\begin{equation}\label{6}
{W^{(t+1)}}={W^{(t)}}-{\alpha}{{\mathbb{E}}_{{(x,y^{'})}{\in}D^{'}}}{{\frac{{\partial}L^{(t+1)}}{{\partial}W}}\bigg|_{W^{(t)}}}
\end{equation}
\begin{equation}\label{7}
{L^{(t+1)}}={\mathcal{L}}({{f_W^{(t)}}(x)},{g_{\theta}^{(t+1)}(h(x),y^{'})})
\end{equation}

\section{EVALUATION}
\subsection{Datasets}
In the experiment, we selected two datasets MoNuSeg \cite{kumar2019multi} and ccRCC to verify our method's superiority. MoNuSeg contains 30 images with pixel-level annotations, each image with a size of 1000×1000 pixels. The ccRCC dataset is constructed and annotated by our team. The original whole slide image is downloaded from TCGA \cite{tomczak2015cancer}. Experienced pathologists select the appropriate area and use the OpenHI \cite{puttapirat2018openhi} platform for annotation. ccRCC dataset contains 1000 diagnostic areas with 512×512 pixels.\par

Based on ccRCC and MoNuSeg datasets, we utilize two methods to generate three datasets (S-ccRCC, S-MoNuSeg, M-ccRCC) used in our experiment. Two kinds of generation methods are as flows:
\subsubsection{Single-nuclei patch generation}
we crop each nucleus according to the pixel-level annotation to ensure that the nuclei are in the center of the patch, then scale all patches to a size of 64×64 pixels. 
\subsubsection{Multi-nuclei patch generation}
we use the sliding window method to crop 128×128 areas from the original image and fine-grained annotations. Each area contains multiple nuclei and corresponding pixel-level annotations. \par
We use the single-nuclei generation method to process ccRCC and MoNuSeg datasets and generate two datasets,i.e., S-ccRCC and S-MoNuSeg. Then Multi-nuclei patch generation method is used to process the ccRCC dataset and generate M-ccRCC. The specific dataset statistics are shown in Table \ref{t1}. \par

We use Iou and Dice coefficients as the evaluation indicators of the experimental results. The calculation formula of Iou and Dice is shown in Eq. (8) and Eq. (9), in which P represents model prediction result and G represents corresponding ground truth mask.\par
\begin{equation}\label{...}
Iou = \frac{P {\cap} G}{P {\cup} G}
\end{equation}
\begin{equation}\label{...}
    Dice = \frac{2 {\times} P {\cap} G}{P + G}
\end{equation}

\begin{table}[h!]
\centering
\caption{Dataset statistics}
\label{t1}
\begin{tabular}{c|c|c|c}
\hline
Methods    & \multicolumn{2}{c|}{Single-nuclei patch} & Multi-nuclei patch \\ \hline
Dataset & S-ccRCC           & S-MoNuSeg          & M-ccRCC        \\ \hline
Train   & 2000            & 1900             & 1000         \\ 
Test    & 1500            & 1200             & 800          \\ 
Meta    & 50              & 30               & 50           \\ \hline
\end{tabular}
\end{table}

\begin{figure}[!t]
\centering
\includegraphics[width=85mm]{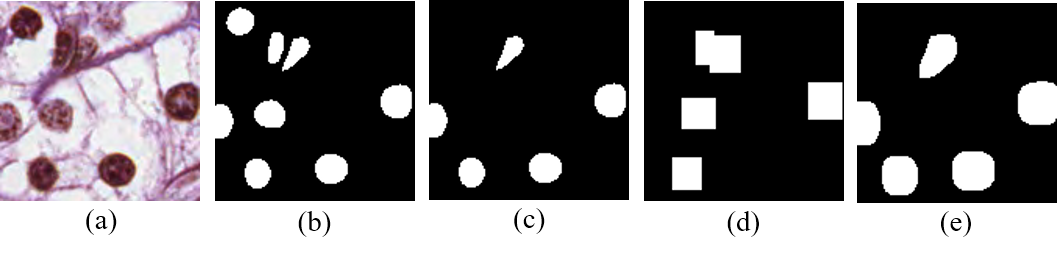}
\caption{Different kinds of annotation. (a) Original Image, (b) Pixel-level annotation, (c) Partial annotation, (d) Bounding box annotation, (e) Dilation annotation.}
\label{f4}
\end{figure}

\subsection{Noisy mask generation}
Fine-grained pixel-level annotation of the whole histopathological image usually takes time and effort. The annotator needs to outline each nucleus and checks repeatedly to confirm whether any nuclei are missed in the current area. Partial coarse-grained annotation can greatly reduce the workload of annotators and improve the efficiency of annotating. In order to simulate the real scene encountered by pathologists and annotators in the annotating process, we set the following ways to add noise to the mask:
\subsubsection{Partial gold annotation}
Based on a certain proportion, we randomly delete a part of the nuclei's annotation and keep other gold annotations as shown in Fig. \ref{f4} (c). 
\subsubsection{Partial weak annotation}
We use two methods to generate different weak annotations for all datasets, i.e., bounding box, dilation noise as shown in Fig. \ref{f4} (d) and (e). We first use the Partial gold annotation method to process the dataset. On this basis, for bounding box noise, each nuclei's mask is expanded to a corresponding circumscribed rectangle, and then the rectangle is randomly expanded by 1 to 3 pixels. For dilation noise, the dilation morphology operator is employed to extend the foreground region by 1 to 5 pixels.

\subsection{Implementation details}
We utilize the PyTorch framework to implement our MMC method. The architecture of ResNet-32 \cite{he2016deep} backbone U-Net \cite{ronneberger2015u} is the main segmentation model. The meta-model structure includes a layer of 3×3 and 1×1 convolution. The learning rate of the main segmentation model is set to  $1 {\times} 10^{-3}$ and drops 0.1 after 300 epochs, and a total of 500 epochs are trained. The learning rate of meta-model is set to $1 {\times} 10^{-4}$, and Adam \cite{kingma2014adam} is used to optimize the two network parameters.

\begin{table*}[h!]
\caption{Comparison results under different noisy types}
\centering
\label{t2}
\begin{tabular}{c|c|c|c|c|c|c|cccccc}
\hline
Dataset & \multicolumn{6}{c|}{S-ccRCC} & \multicolumn{6}{c}{S-MoNuSeg} \\ \hline
Noise & \multicolumn{2}{c|}{Bounding box} & \multicolumn{2}{c|}{Dilation} & \multicolumn{2}{c|}{Partial} & \multicolumn{2}{c|}{Bounding box} & \multicolumn{2}{c|}{Dilation} & \multicolumn{2}{c}{Partial} \\ \hline
Metric & Dice & Iou & Dice & Iou & Dice & Iou & \multicolumn{1}{c|}{Dice} & \multicolumn{1}{c|}{Iou} & \multicolumn{1}{c|}{Dice} & \multicolumn{1}{c|}{Iou} & \multicolumn{1}{c|}{Dice} & Iou \\ \hline
U-net FT & 0.5423 & 0.3747 & 0.6349 & 0.4683 & 0.6133 & 0.4454 & \multicolumn{1}{c|}{0.5371} & \multicolumn{1}{c|}{0.3926} & \multicolumn{1}{c|}{0.6537} & \multicolumn{1}{c|}{0.5068} & \multicolumn{1}{c|}{0.6687} & 0.5315 \\
SR & 0.6185 & 0.4159 & 0.7079 & 0.5534 & 0.6256 & 0.4763 & \multicolumn{1}{c|}{0.6281} & \multicolumn{1}{c|}{0.4611} & \multicolumn{1}{c|}{0.6820} & \multicolumn{1}{c|}{0.5199} & \multicolumn{1}{c|}{0.6726} & 0.5572 \\
MCPM & 0.7016 & 0.5426 & 0.7449 & 0.5955 & 0.7497 & 0.6014 & \multicolumn{1}{c|}{0.6172} & \multicolumn{1}{c|}{0.4472} & \multicolumn{1}{c|}{0.7483} & \multicolumn{1}{c|}{0.5991} & \multicolumn{1}{c|}{0.7196} & 0.5633 \\
MMC & 0.8395 & 0.7244 & \textbf{0.8642} & \textbf{0.7619} & \textbf{0.8696} & \textbf{0.7700} & \multicolumn{1}{c|}{0.6887} & \multicolumn{1}{c|}{0.5271} & \multicolumn{1}{c|}{0.7837} & \multicolumn{1}{c|}{0.6449} & \multicolumn{1}{c|}{0.7882} & 0.6509 \\
U-net Clean & \textbf{0.8599} & \textbf{0.7557} & 0.8599 & 0.7557 & 0.8599 & 0.7557 & \multicolumn{1}{c|}{\textbf{0.7896}} & \multicolumn{1}{c|}{\textbf{0.6537}} & \multicolumn{1}{c|}{\textbf{0.7896}} & \multicolumn{1}{c|}{\textbf{0.6537}} & \multicolumn{1}{c|}{\textbf{0.7896}} & \textbf{0.6537} \\ \hline
\end{tabular}
\end{table*}

\subsection{Comparison with Existing Methods}
\begin{figure}[!t]
\centering
\includegraphics[width=60mm]{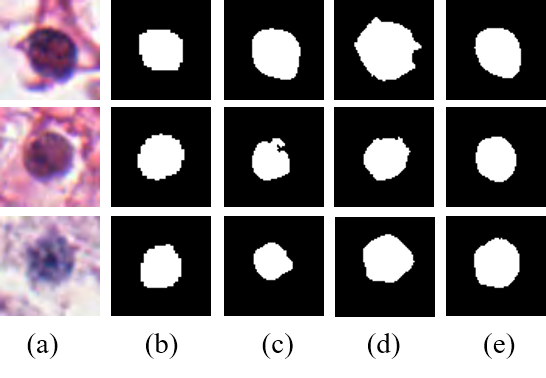}
\caption{S-ccRCC segmentation result. (a) original image, (b) pixel-level ground truth, (c) spatial reweighting, (d) MCPM, (e) MMC.}
\label{f2}
\end{figure}

We conducted experiments on both single- and multi-nuclei patch datasets. In the single-nuclei patch segmentation, we conducted comparative experiments under all three kinds of noise. The experimental results are shown in Table \ref{t2}. 'U-Net FT' means using noisy data for training and then fine-tuning on the clean meta-data. 'U-Net Clean' represents optimal performance that the segmentation model can achieve if there is no noise in the supervised scenario's training data. We chose two latest medical image segmentation methods based on meta-learning, MCPM and spatial reweighting(termed SR) for comparison. According to the above-mentioned noisy generation method, we add ${40\%}$ noise for each condition. Our method achieves the best results under both the Dice and Iou metrics at S-ccRCC. For bounding box, dilation, and partial gold noise, our method's Dice value has achieved performance improvements of 0.1397, 0.1194, and 0.1199 respectively.  Our method even achieved better results than training with clean data in the supervised scenario for dilation and partial gold noise. In bounding box noise, the performance gap between our method and 'U-Net supervised' is just about 2 percent. Our method also achieves the best result at the S-MoNuSeg dataset. In the partial gold noise condition, the segmentation result is shown in Fig. \ref{f2}. The contour of nuclei segmented by our method is more accurate. This shows that our method can better suppress the influence of noise labels in the training data under the condition of only a small amount of fine-grained annotations and can train a model with satisfactory segmentation performance. \par
\begin{figure}[!t]
\centering
\includegraphics[width=80mm]{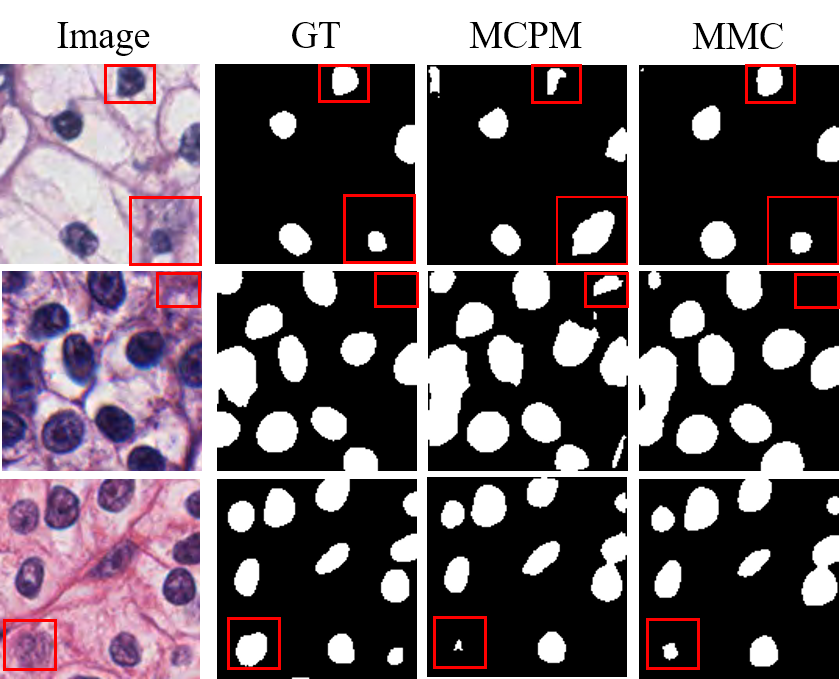}
\caption{M-ccRCC segmentation result. Red box marked places show the advantages of MMC.}
\label{f3}
\end{figure}

In the multi-nuclei patch segmentation experiment, our method still achieved the best performance under three kinds of noisy types, as shown in Table \ref{t3}.
For bounding box,  dilation, and partial gold noise, our method's Dice value has achieved performance improvements of 0.0494, 0.0925, and 0.0572, respectively. The segmentation results of the multi-nuclei patch are shown in Fig. \ref{f3}. Our method can better suppress the influence of noise, and has a higher recognition ability for foreground and background.\par
We also verified the influence of different meta-model structures on the segmentation performance. From the Table \ref{t4}, it can be seen that in the condition of partial gold noise, when the meta-model structure has two convolutional layers with kernel in size of 3×3 and 1×1, our method achieves the best performance.

\begin{table}[h!]
\centering
\caption{M-ccRCC experiment results}
\label{t3}
\begin{tabular}{cccc}
\hline
M-ccRCC & Bounding box & Dilation & Partial \\ \hline
MCPM & 0.6732 & 0.6780 & 0.6967 \\
MMC & \textbf{0.7228} & \textbf{0.7705} & \textbf{0.7539} \\ \hline
\end{tabular}
\end{table}

\begin{table}[h!]
\centering
\caption{The influence of meta-model structures}
\label{t4}
\begin{tabular}{ccc}
\hline
Meta-model & S-ccRCC & S-MoNuSeg \\ \hline
3*3 + 1*1 & \textbf{0.8696} & \textbf{0.8069} \\ 
3*3 + 3*3 + 1*1 & 0.8114 & 0.7296 \\ 
3*3 + 5*5 + 1*1 & 0.7464 & 0.7277 \\ \hline
\end{tabular}
\end{table}

\section{CONCLUSION}
This paper proposes a novel meta-learning method MMC to learn an automated nuclei segmentation model with noisy annotation. Only a very small amount of high-quality and pixel-level annotation is needed, and the meta-model can map a large number of noisy masks to corrected masks, which supervises the main segmentation model training to achieve good performance. The parameters of the main segmentation model and meta-model adopt a bi-level optimization strategy. Extensive experimental results on two datasets show that MMC has superior performance compared with the latest loss-weighted methods. It even exceeds the performance of the model trained on clean data in the supervised scenario in some settings.

\section{DISCUSSION}
The proposed method in this paper solves how to suppress the influence of noise in the nuclei segmentation problem and achieves the best results. However, the bi-level optimization method adopted by MMC requires three times the training time compared to the traditional training method. The paper uses ResNet-32 for feature extraction during the experiment. When the backbone is replaced with a structure with stronger feature extraction capability, the performance of MMC can be further improved. The proposed method is also universal, and the segmentation model can be easily replaced with any structure.

\section{Acknowledgment}
This work has been supported by National Natural Science Foundation of China (61772409); The consulting research project of the Chinese Academy of Engineering (The Online and Offline Mixed Educational Service System for “The Belt and Road” Training in MOOC China); Project of China Knowledge Centre for Engineering Science and Technology; The innovation team from the Ministry of Education (IRT\_17R86); and the Innovative Research Group of the National Natural Science Foundation of China (61721002). The results shown here are in whole or part based upon data generated by the TCGA Research Network: https://www.cancer.gov/tcga.



\end{document}